\documentclass[letterpaper, 10 pt, conference]{ieeeconf}  

\IEEEoverridecommandlockouts                              

\usepackage{graphics} 
\usepackage{epsfig} 
\usepackage{times} 
\usepackage{amsmath} 
\usepackage{amssymb}  
\usepackage{booktabs}
\usepackage{multirow}
\usepackage[table,xcdraw]{xcolor}
\graphicspath{ {./images/} }
\usepackage{flushend}
\usepackage{cite}
\usepackage{bm}
\usepackage{hyperref}
\hypersetup{hidelinks,
colorlinks=true,
allcolors=black,
pdfstartview=Fit,
breaklinks=true}
\usepackage{color}
\usepackage{siunitx} 
\usepackage{ulem}

\title{\LARGE
Learning Multiple Gaits within Latent Space for Quadruped Robots
}

\author{Jinze Wu, Yufei Xue, Chenkun Qi$^{*}$
\thanks{All authors are with School of Mechanical Engineering, Shanghai Jiao Tong University, Shanghai, China. (Email: wjz\_sjtu@sjtu.edu.cn; xue\_yeiii@sjtu.edu.cn; chenkqi@sjtu.edu.cn;).}
\thanks{This work was supported by the National Key Research and Development Plan (2021YFF0307900), the Fundamental Research Funds for the Central Universities (82232025).}
\thanks{*Corresponding author}
}

\begin{document}

\maketitle

\thispagestyle{empty}
\pagestyle{empty}

\begin{abstract}

Learning multiple gaits is non-trivial for legged robots, especially when encountering different terrains and velocity commands. In this work, we present an end-to-end training framework for learning multiple gaits for quadruped robots, tailored to the needs of robust locomotion, agile locomotion, and user's commands. A latent space is constructed concurrently by a gait encoder and a gait generator, which helps the agent to reuse multiple gait skills to achieve adaptive gait behaviors. To learn natural behaviors for multiple gaits, we design gait-dependent rewards that are constructed explicitly from gait parameters and implicitly from conditional adversarial motion priors (CAMP). We demonstrate such multiple gaits control on a quadruped robot Go1 with only proprioceptive sensors.

\end{abstract}

\section{INTRODUCTION}
Animals have evolved a spectrum of locomotion gaits to maximize their robustness and efficiency at different terrains and speeds \cite{hoyt1981gait,alexander1984gaits}. 
Reproducing such natural gait transitions has been a challenging topic in the legged robotic community due to the complexity of multi-legged systems. 
To produce stable gait motions, most methods use model predictive control (MPC) to design locomotion controllers \cite{MITCheetah3,NLPoptimizationQuadruped,ding2021representation}. These MPC methods often require online optimization with simplified dynamic models and are fundamentally restricted to predefined contact sequences on quasi-flat grounds. 
While some frameworks \cite{lengagne2013generation,towrNote} that can optimize contact sequences and leg movements for complex scenarios have also been developed, they are too computationally intensive to be used in real-time. 
One compromise is to decompose the locomotion problem into two different sub-problems: contact planning and motion optimization \cite{ponton2021efficient,meduri2023biconmp}. Although this decomposition reduces the complexity of the overall problem, it requires degrees of expert knowledge to solve individual sub-problems and does not guarantee the optimality of the final result.

Recently, a large number of works have designed robust and agile locomotion controllers for legged robots using reinforcement learning (RL) \cite{hwangbo2019learning,lee2020learning,kumar2021rma,miki2022learning,margolis2022rapid,ji2022concurrent,rudin2022learning,wjzamp}. These RL-based controllers have shown strong performance and generalization both indoors and outdoors. Their success is mainly due to the following two properties. 
One is that RL provides a uniform framework to learn a control policy from the agent's interactions with the environment, which facilitates the design of robust locomotion controllers that take into account both the robot and the environment \cite{lee2020learning,kumar2021rma,miki2022learning}. 
The other is that the policy network learned from RL provides a direct mapping from observations to actions, which is more likely to achieve real-time control of agile movements \cite{margolis2022rapid,ji2022concurrent}.  
However, many works \cite{kumar2021rma,margolis2022rapid,ji2022concurrent} employing RL methods exhibit unnatural and jerky legged movements, since it is difficult to achieve desired behaviors without well-designed reward functions in the RL domain. 
Moreover, the main positive rewards for training locomotion in most RL methods come from the tracking terms of velocity commands \cite{hwangbo2019learning,lee2020learning,kumar2021rma,miki2022learning,margolis2022rapid,ji2022concurrent,rudin2022learning,wjzamp}, which makes most RL controllers only capable of generating specific gait movements without user control over gait details.

\subsection{Related RL Methods for Locomotion}
Adding predefined gait priors into training procedure may accelerate the convergence to a desired gait for legged robots. One approach is to use imitation learning to mimic the reference motion of a specific expert by explicitly using
a phase variable \cite{peng2018deepmimic,peng2020learning}. 
Although this technique can speed up training, it is difficult to scale to multiple reference motions.  
Another approach is to construct a specific action space for the agent, such as central pattern generators (CPGs) \cite{shi2022reinforcement,2022PhaseGuidedController} or policies modulate trajectory generators (PMTG) \cite{2018pmtg,lee2020learning}. However, these specific action spaces have large limitations on the actions of the agent and are difficult to produce complex behaviors.

In order to learn user-specified gait behaviors without limiting the ability of robust and agile locomotion, adversarial motion priors (AMP) is a compromise approach \cite{escontrela2022adversarial,li2022versatile,wjzamp}. Compared to direct imitation learning, AMP uses style rewards to guide the agent to produce desired motions, which provides flexibility for learning complex behaviors with natural styles. 
However, AMP is prone to mode-collapse when applied to the dataset consisting of multiple behaviors, where the agent only learns a narrow range of behaviors in the original dataset.

To learn multiple gaits, contact-related rewards can be used by integrating desired contact patterns into the observation space and composing probabilistic periodic costs over contact forces and velocities  \cite{2021BipedalPeriodicReward,margolis2022walktheseways}. 
However, during the training of multi-task RL, the robot needs to coordinate the tracking ability of velocity commands and the detailed control ability of gait behaviors. This may cause the robot to fail to learn the gait behaviors that can adapt to the external environments and its own speeds, thus sacrificing the ability of robust or agile locomotion.

To avoid task conflict in multi-task RL, hierarchical policies and latent space models can be used to learn reusable skills from large motion datasets by explicitly dividing the RL problem into pre-training and transfer phase \cite{peng2019mcp,2022-TOG-ASE}.
With this two-phase split, the agent can focus on skill learning in the pre-training phase and on how to use the skills learned in the first phase to complete a specific task in the task learning phase. 
However, most of these hierarchical frameworks are trained on flat ground, freezing the low-level skill network in the second phase. This freeze may cause the low-level skill network to fail to generalize to complex terrains that are not encountered in the first phase.

\begin{figure*}[htbp]
\setlength{\abovecaptionskip}{0.cm}
\setlength{\belowcaptionskip}{-0.cm}
\centering
\includegraphics[width=\linewidth]{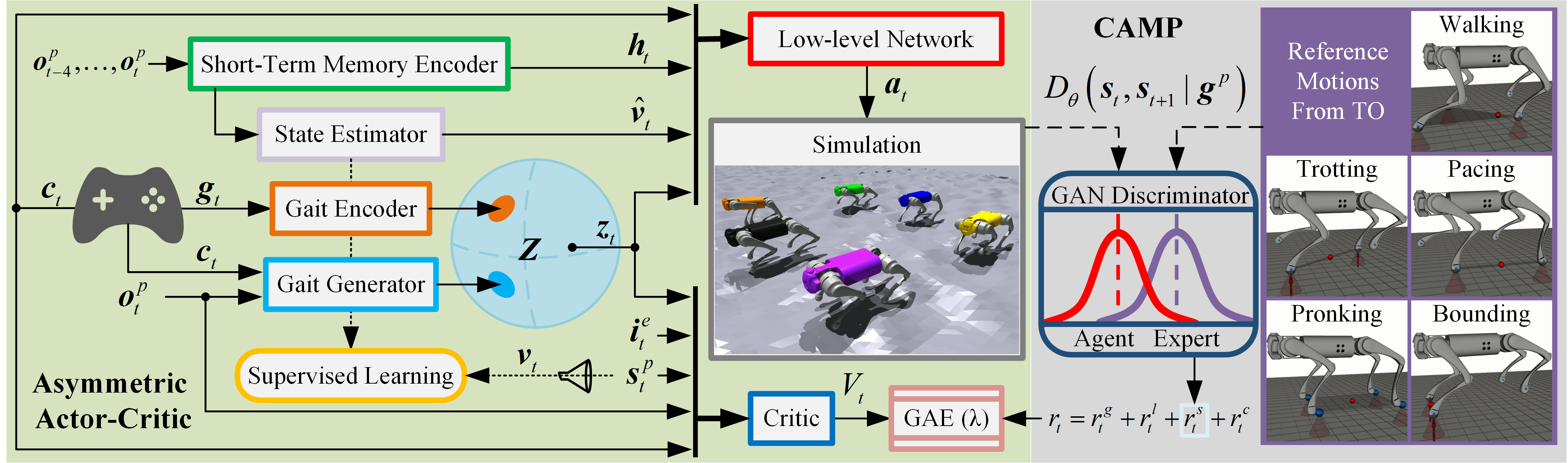}
\caption{Overview of the learning framework. We employ the asymmetric actor-critic framework to learn robust blind locomotion in one training phase. A latent space $\boldsymbol{Z}$ is constructed concurrently by a gait encoder and a gait generator, which helps the agent to reuse multiple gait skills to achieve adaptive gait behaviors. We use a conditional discriminator to guide the policy to reproduce the natural gait motions that are similar to the CAMP dataset generated from TO.}
\label{fig:method}
\end{figure*}

\subsection{Contributions}
To further improve the controllability of RL controllers for legged robots, we propose an end-to-end training framework for learning multiple gaits for quadruped robots. 
Our controller can control the gait behaviors in detail under user's commands and achieve smooth transitions of various gaits consisting of walking, trotting, pacing, pronking, and bounding. 
Moreover, our controller can reuse learned gait skills to generate adaptive gait behaviors to achieve robust and agile locomotion.
The main contributions are listed as follows:

\begin{enumerate}
\item We construct the latent space of gait skills by concurrently using a gait encoder and a gait generator, which helps our robot to reuse multiple gait skills to achieve adaptive gait behaviors.
\item We design gait-dependent rewards to guide our robot to learn natural gait skills in the latent space. The gait-dependent rewards are constructed explicitly from gait parameters and implicitly from conditional adversarial motion priors (CAMP).
\item Outdoor experiments show that our robot can achieve multiple gaits according to user's commands, while achieving robust and agile locomotion.
\end{enumerate}

\section{METHOD}
\subsection{Reinforcement Learning Problem Formulation}
Since the terrains are not fully observable without exteroceptive sensors, our locomotion problem is modeled as a partially observable Markov decision process (POMDP). The environment is completely defined by a full state $\boldsymbol{x}_t$ at time step $t$. The agent's policy performs an action $\boldsymbol{a}_t$, and then the environment moves to the next state $\boldsymbol{x}_{t+1}$ with a transition probability \(P\left({{\boldsymbol{x}_{t + 1}}\mid{\boldsymbol{x}_t},{\boldsymbol{a}_t}}\right)\) and returns a reward $r_t$ and a partial observation $\boldsymbol{x}_{t+1}^p$. The goal of RL is to find a policy $\pi$ to maximize the expected discounted return over the future trajectory:
\begin{equation} \label{RL def}
J\left(\pi\right) = {\mathbb{E}_{{\pi}}}\left[ {\sum\limits_{t = 0}^\infty  {{\gamma ^t}{r_t}} } \right],
\end{equation}
where $\gamma ^t\in\left[0, 1\right)$ is a discount factor.

Recent works have leveraged the teacher-student training paradigm to address the POMDP \cite{lee2020learning,kumar2021rma,miki2022learning,wjzamp}. 
While it has been empirically shown that the student policy can achieve similar performance to the teacher policy, the student can never outperform the teacher. 
Moreover, training the teacher and student networks sequentially requires more data, which is sample inefficient.

To learn robust blind locomotion in one training phase, we train our policy using proximal policy optimization (PPO) \cite{schulman2017proximal}, employing the asymmetric actor-critic framework used in \cite{Pinto2018AsymmetricAC}. 

\textbf{State Space:} 
The critic is trained on the full state consisting of partial observation $\boldsymbol{o}_t^p$,  velocity command $\boldsymbol{c}_t$, privileged state $\boldsymbol{s}_t^p$, terrain information $\boldsymbol{i}_t^e$, and latent representation $\boldsymbol{z}_t$ of gait skills, while the actor has access to the same information except for $\boldsymbol{s}_t^p$ and $\boldsymbol{i}_t^e$. This asymmetric framework allows us to train policy networks that can be easily deployed on the real robot with onboard sensors, while exploiting the availability of the full state in physics simulators. 

The partial observation $\boldsymbol{o}_t^p\in {\mathbb{R}^{42}}$ contains the orientation of the gravity vector in the robot's base frame, base angular velocity, joint positions and velocities, and the previous action $\boldsymbol{a}_{t-1}$ selected by the current policy.
The velocity command $\boldsymbol{c}_t=\left(v_x^{\rm cmd},v_y^{\rm cmd},\omega_z^{\rm cmd}\right)\in {\mathbb{R}^{3}}$ represents the desired longitudinal, lateral, and yaw velocities in the robot's base frame.
The privileged state $\boldsymbol{s}_t^p\in{\mathbb{R}^{21}}$ consists of base linear velocity, contact forces, external forces and their positions on the robot. 
The terrain information $\boldsymbol{i}_t^e\in{\mathbb{R}^{187}}$ represents the height map scan of the robot's surroundings, containing 187 points sampled from a grid around the robot's base.
The latent representation $\boldsymbol{z}_t\in{\mathbb{R}^{16}}$ of gait skills is computed from the gait encoder or the gait generator, which guides the low-level network to produce specific gaits or adaptive gait behaviors. More details about the $\boldsymbol{z}_t$ are given in Section \ref{subsection: latent space}.

\textbf{Action Space:} The policy outputs 12 joint position offsets as the action $\boldsymbol{a}_{t}\in{\mathbb{R}^{12}}$. The desired joint positions are equal to the joint position offsets plus the time-invariant nominal joint positions and are sent to low-level joint PD controllers with fixed gains ($K_p=20$, $K_d=0.5$).

\subsection{Network Architecture} \label{subsection: network architecture}
The overall architecture of our learning framework is shown in Fig. \ref{fig:method}. Both of the actor and critic are neural networks designed as the Multilayer Perceptron (MLP) with ELU activations for hidden layers. The actor contains five MLP parts: a short-term memory encoder (STMEnc), a state estimator, a gait encoder, a gait generator, and a low-level network. More details on each layer are shown in Table \ref{tab:net_arc}.

\begin{table}[htbp]
\setlength{\abovecaptionskip}{0.cm}
\setlength{\belowcaptionskip}{-0.cm}
\centering
\caption{Network architectures}
\label{tab:net_arc}
\begin{tabular}{@{}lllll@{}}
\toprule
\textbf{Module} & \textbf{Inputs} & \textbf{Hidden Layers} & \textbf{Outputs} \\ \midrule
STMEnc &$\boldsymbol{o}_{t-4}^p,...,\boldsymbol{o}_{t-1}^p,\boldsymbol{o}_{t}^p$&{[}256, 128{]}& $\boldsymbol{h}_t$ \\
State Estimator& $\boldsymbol{h}_t$    & {[}64, 32{]}            & $\hat{\boldsymbol{v}}_t$  \\
Gait Encoder& $\boldsymbol{g}_{t}$   & {[}64, 32{]}  & $\bar{\boldsymbol{z}}_t$ \\
Gait Generator& $\boldsymbol{c}_t,\boldsymbol{o}_{t}^p$ & {[}128, 64{]}  & $\bar{\boldsymbol{z}}_t$ \\
Low-Level &$\boldsymbol{c}_t,\boldsymbol{h}_t,\hat{\boldsymbol{v}}_t,\boldsymbol{z}_t$ & {[}256, 128, 64{]}     & $\boldsymbol{a}_t$  \\
Critic&$\boldsymbol{c}_t,\boldsymbol{o}_{t}^p,\boldsymbol{s}_t^p,\boldsymbol{i}_t^e,\boldsymbol{z}_t$&{[}512, 256, 128{]}& $V_t$  \\
$D_{\theta}$& $\boldsymbol{s}_t^{CAMP},\boldsymbol{s}_{t+1}^{CAMP},\boldsymbol{g}^{p}$ & {[}1024, 512{]}& $d_t^{\rm score}$ \\ \bottomrule
\end{tabular}
\end{table}

Since the agent in the POMDP needs to consider the histories of the partial observation $\boldsymbol{o}_t^p$ to select an action, we first design the STMEnc for our robot to encode the sequential correlations between short-term histories. The STMEnc receives the five most recent partial observations $\boldsymbol{o}_{t-4}^p,...,\boldsymbol{o}_{t}^p$ as input and outputs an encoded history $\boldsymbol{h}_t\in{\mathbb{R}^{32}}$, which is essential for our robot to estimate the privileged state and terrain information implicitly. Motivated by \cite{ji2022concurrent}, we concurrently train the state estimator with other networks to explicitly estimate base linear velocity $\hat{\boldsymbol{v}}_t$ using supervised learning. 
The $\hat{\boldsymbol{v}}_t$ can help the low-level network to understand how well the locomotion task is performed under the current $\boldsymbol{c}_t$, while the $\boldsymbol{h}_t$ can help the low-level network to be aware of the interactions between the robot and the environment.
The $\boldsymbol{c}_t$, $\boldsymbol{h}_t$, $\hat{\boldsymbol{v}}_t$, and $\boldsymbol{z}_t$ are then concatenated and fed to the low-level network with a $tanh$ output layer to output the mean $\boldsymbol{\mu}_t \in {\mathbb{R}^{12}}$ of a Gaussian distribution $\boldsymbol{a}_t\sim\mathcal{N}\left(\boldsymbol{\mu}_t,\boldsymbol{\sigma} \right)$, where $\boldsymbol{\sigma} \in {\mathbb{R}^{12}}$ denotes the variance of the action determined by PPO. 

\begin{figure}[htbp]
\setlength{\abovecaptionskip}{0.cm}
\setlength{\belowcaptionskip}{-0.cm}
\centering
\includegraphics[width=\linewidth]{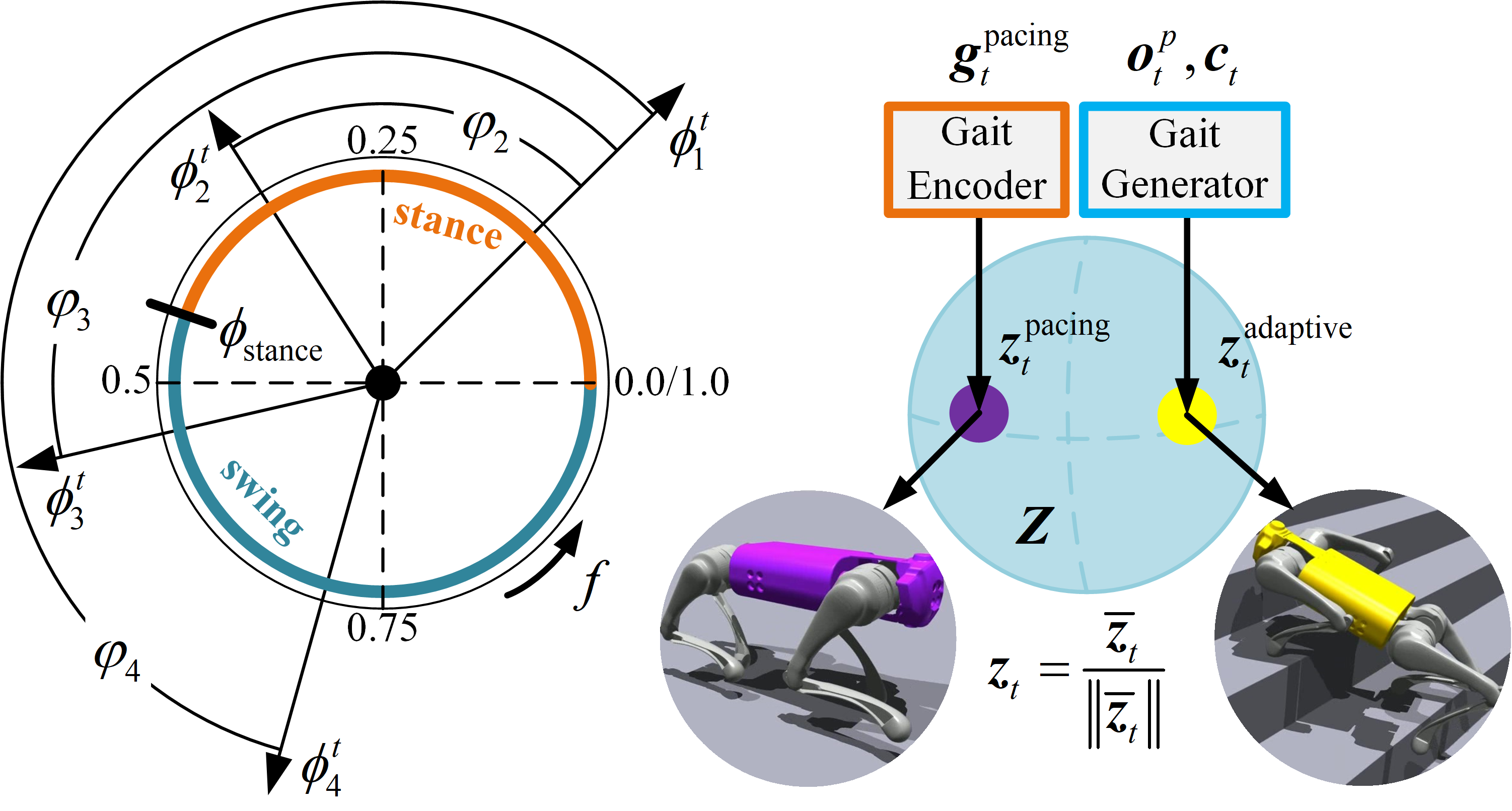}
\caption{The original space of the gait parameters (left) and the latent space of the gait skills (right).}
\label{fig:z_space}
\end{figure}

\subsection{Latent Space for Gait Skills} \label{subsection: latent space}
To represent multiple gaits for quadruped robots, we first design the original space of gait parameters $\boldsymbol{g}_t$ for our robot. 
Motivated by previous works \cite{2021BipedalPeriodicReward,margolis2022walktheseways}, we represent the locomotion gait of quadruped robots by four independent periodic phases ${\phi}_i\in\left[0,1\right),i=1,2,3,4$. 
As shown in Fig. \ref{fig:z_space}, each phase variable ${\phi}_i$ varies from 0 to 1, wraps back to 0, and starts a new cycle. 
Each leg starts with the stance status (${\phi}_i=0$) at the beginning of a gait cycle, increases monotonically with a stepping frequency ${f}$, and switches to the swing status after reaching a threshold (${\phi}_i > {\phi}_{\rm stance}$). The stance ratio ${\phi}_{\rm stance}\in\left(0,1\right)$ is the proportion of  the standing time in each gait cycle, which controls the switching point for swing. 
To reduce the dimension of the original space of gait parameters, we use the same ${f}$ and ${\phi}_{\rm stance}$ across all legs, and allow the last three legs (front-right, rear-left, rear-right) to have different phase offsets ${\varphi}_i\in\left(0,1\right),i=2,3,4$ compared to the front-left leg. 
Using only three phase offsets ${\varphi}_{2,3,4}$, common gaits including walking $\left(0.5,0.25,0.75\right)$, trotting $\left(0.5,0.5,0\right)$, pacing $\left(0.5,0,0.5\right)$, pronking $\left(0,0,0\right)$, and bounding $\left(0,0.5,0.5\right)$ can be represented.
Specifically, at each control step ($dt=\SI{0.02}{s}$), the phase of each individual leg is computed by:
\begin{equation} \label{leg phase compute}
\begin{aligned}
&{\phi}_1^{t+1} = \left({\phi}_1^{t}+ f\times dt\right)\left(mod\:1\right)\\
&{\phi}_i^{t+1} = \left({\phi}_1^{t+1}+{\varphi}_i\right)\left(mod\:1\right),i=2,3,4.
\end{aligned}
\end{equation}
To improve the traversability of different gaits when encountering obstacles in the altitude direction, we provide the target base height ${h}_{b}^{\rm cmd}$ in the original space of the $\boldsymbol{g}_t$ . Specifically, the $\boldsymbol{g}_t\in {\mathbb{R}^{8}}$ is defined as:
\begin{equation} \label{gait parameters}
\left[ sin(2\pi{\phi}_1^{t}),cos(2\pi{\phi}_1^{t}),{\varphi}_2,{\varphi}_3,{\varphi}_4,f,{\phi}_{\rm stance},h_b^{\rm cmd} \right],
\end{equation}
which is expressive enough to represent a rich set of locomotion gaits.

To learn multiple gaits for our robot, we construct a latent space for learning reusable gait skills and producing adaptive gait behaviors. 
An overview of this latent space is provided in Fig. \ref{fig:z_space}.  
We first encode the $\boldsymbol{g}_t$ into an unbounded latent representation $\bar{\boldsymbol{z}}_t$ using the gait encoder. 
Inspired by prior works \cite{peng2022ase,tessler2023calm}, the $\bar{\boldsymbol{z}}_t$ is projected onto the $l_2$ unit hypersphere to produce a bounded latent space of $\boldsymbol{z}_t$. This $l_2$ norm constraint reduces the likelihood of unnatural behaviors caused by $\boldsymbol{z}_t$ outside the distribution encoded from the $\boldsymbol{g}_t$ during inference, which helps the agent to learn reusable gait skills for producing adaptive gait behaviors. 
Additionally, we train the gait generator to construct the same latent space by encoding the current states ($\boldsymbol{o}_t^p$, $\boldsymbol{c}_t$). This gait generator can exploit the bounded latent space to reuse learned gait skills to generate adaptive behaviors for robust or agile locomotion, based on the current feedback states.

To ensure the stability of the training, the latent space can only be constructed by one of the gait encoder or the gait generator at each training step. However, it is easy to construct the latent space concurrently by the gait encoder and the gait generator with massively parallel agents \cite{rudin2022learning}. 
More details about the training procedure are given in Section \ref{subsection: Gait Curriculum}

\subsection{Reward Functions Design}\label{subsection: Reward Terms Design}
Learning desired behaviors for agents often requires degrees of expert knowledge to design suitable reward functions, which is one of the long-standing challenges in the RL domain. 
Our previous work \cite{wjzamp} has shown that using AMP can help robots to learn natural gait without limiting their ability to overcome challenging terrains. However, this approach is difficult to scale to learn multiple gaits since the AMP is prone to mode-collapse when applied to dataset consisting of multiple behaviors.
Motivated by \cite{tessler2023calm}, we help our robot to learn multiple gaits by using a conditional discriminator that guides the policy to reproduce the natural gait motions that are similar to the CAMP dataset, which are conditioned on partial gait parameters $\boldsymbol{g}^{p}\in {\mathbb{R}^{5}}$. The $\boldsymbol{g}^{p}$ contains ${\varphi}_{2,3,4},{\phi}_{\rm stance},f$, which is time-invariant for the corresponding periodic gait. 

We first construct the motion dataset of CAMP conditioned on different $\boldsymbol{g}^{p}$, applying the same trajectory optimization (TO) technique used in our previous work \cite{wjzamp}. The motion dataset of CAMP is detailed in Table \ref{tab:CAMP dataset}, containing five common gaits of walking, trotting, pacing, pronking, and bounding. Each gait consists of two locomotion trajectories with slow (\SI{2}{\hertz}) and fast (\SI{4}{\hertz}) frequencies. Each trajectory lasts 2 seconds.

\begin{table}[htbp]
\setlength{\abovecaptionskip}{0.cm}
\setlength{\belowcaptionskip}{-0.cm}
\centering
\caption{Motion dataset of CAMP conditioned on the partial gait parameters.}
\label{tab:CAMP dataset}
\begin{tabular}{@{}llll@{}}
\toprule
Gait name                 & ${\varphi}_{2,3,4}$             & ${\phi}_{\rm stance}$     & ${f}$ [Hz] \\ \midrule
\multirow{2}{*}{Walking}  & \multirow{2}{*}{0.5, 0.25, 0.75}& \multirow{2}{*}{0.75} & 2             \\
                          &                                 &                       & 4             \\ \midrule
\multirow{2}{*}{Trotting} & \multirow{2}{*}{0.5, 0.5, 0}    & \multirow{2}{*}{0.5}  & 2             \\
                          &                                 &                       & 4             \\ \midrule
\multirow{2}{*}{Pacing}   & \multirow{2}{*}{0.5, 0, 0.5}    & \multirow{2}{*}{0.5}  & 2             \\
                          &                                 &                       & 4             \\ \midrule
\multirow{2}{*}{Pronking} & \multirow{2}{*}{0, 0, 0}        & \multirow{2}{*}{0.5}  & 2             \\
                          &                                 &                       & 4             \\ \midrule
\multirow{2}{*}{Bounding} & \multirow{2}{*}{0, 0.5, 0.5}    & \multirow{2}{*}{0.5}  & 2             \\
                          &                                 &                       & 4             \\
\bottomrule 
\end{tabular}
\end{table}

The conditional discriminator $D_{\theta}$ is represented by a neural network with parameters $\theta$, which predicts whether a state transition \(\left(\boldsymbol{s}_t,\boldsymbol{s}_{t+1}\right)\) conditioned on the $\boldsymbol{g}^{p}$ is a real sample from the dataset $\mathcal{D}$ or a fake sample produced by the agent $\mathcal{A}$. 
Each state $\boldsymbol{s}_t^{CAMP}\in {\mathbb{R}^{30}}$ contains the joint positions, joint velocities, base linear velocity, and base angular velocity.

The training objective for the $D_{\theta}$ is defined as: 
\begin{multline} \label{discriminator}
\mathop{\arg\min} \limits_{\theta} \mathbb{E}_{ \left(\boldsymbol{s}_t,\boldsymbol{s}_{t+1},\boldsymbol{g}^{p}\right) \sim \mathcal{D}} \left[\left(D_{\theta} \left(\boldsymbol{s}_t,\boldsymbol{s}_{t+1}|\boldsymbol{g}^{p}\right) - 1\right)^2\right] \\
+\mathbb{E}_{ \left(\boldsymbol{s}_t,\boldsymbol{s}_{t+1},\boldsymbol{g}^{p}\right) \sim \mathcal{A}} \left[\left(D_{\theta} \left(\boldsymbol{s}_t,\boldsymbol{s}_{t+1}|\boldsymbol{g}^{p}\right) + 1\right)^2\right] \\
+\frac{{{\alpha ^{gp}}}}{2}{\mathbb{E}_{\left( \boldsymbol{s}_t,\boldsymbol{s}_{t+1},\boldsymbol{g}^{p} \right)\sim{\cal D}}}\left[ \left \Vert {{\nabla_\theta}{D_{\theta} }\left( \boldsymbol{s}_t,\boldsymbol{s}_{t+1}|\boldsymbol{g}^{p} \right)} \right\Vert_2 \right],
\end{multline}
where the first two terms are least square GAN formulation, the final term is a regularization which mitigates the discriminator's overfitting on the manifold of real data samples. The $\alpha^{gp}$ is a manually-specified coefficient (we use $\alpha^{gp}=10$). 
The style reward is then defined and scaled to the range $\left[0, 1\right]$ by:
\begin{equation} \label{style reward}
r_t^s = \max \left[0, 1-0.25\times\left(d_t^{\rm score}- 1\right)^2 \right],
\end{equation}
where $d_t^{\rm score}=D_{\theta} \left(\boldsymbol{s}_t,\boldsymbol{s}_{t+1}|\boldsymbol{g}^{p}\right),\left(\boldsymbol{s}_t,\boldsymbol{s}_{t + 1},\boldsymbol{g}^{p}\right)\sim \mathcal{A}$.

Although the style reward $r_t^s$ can help the agent to lean motion style implicitly from the dataset of CAMP, these motion priors are too soft to learn the detailed control of the gait behaviors based on user's commands.
Motivated by \cite{margolis2022walktheseways}, we add a contact-related term $r_t^c$ to the total reward $r_t$ by composing probabilistic periodic costs over contact forces and velocities. At each control step, a desired contact schedule $C_i^{\rm des}\left({\phi}_i\right),i=1,2,3,4$ is computed by \cite{margolis2022walktheseways}:

\begin{multline} \label{contact schedule}
\bar{\phi}_i=\left\{
\begin{array}{lr}
0.5\times\dfrac{{\phi}_i}{{\phi}_{\rm stance}}, &\phi_i\leq{\phi}_{\rm stance}.\\
0.5+0.5\times\dfrac{{\phi}_i-{\phi}_{\rm stance}}{1-{\phi}_{\rm stance}},
&\phi_i > {\phi}_{\rm stance}.\\
\end{array}
\right.\\
\begin{aligned}
C_i^{\rm des}\left({\phi}_i\right)=& \Phi\left(\bar{\phi}_i,\sigma\right)\left[1-\Phi\left(\bar{\phi}_i-0.5,\sigma\right)\right]\\
+&\Phi\left(\bar{\phi}_i-1,\sigma\right)\left[1-\Phi\left(\bar{\phi}_i-1.5,\sigma\right)\right],
\end{aligned}
\end{multline}
where $\Phi\left(x,\sigma\right)$ is the cumulative density function of the normal distribution (we use $\sigma=0.05$).
Each $C_i^{\rm des}\left({\phi}_i\right)$ represents whether the corresponding leg should stand or swing in terms of probability, and is used to compute periodic costs penalizing swing velocities or contact forces by:
\begin{multline} \label{contact-related reward}
\begin{aligned}
r_t^c =& -\sum_{i=1}^{4}\left[1-C_i^{\rm des}\left({\phi}_i\right)\right]\left[1-\exp\left(-\Vert\boldsymbol{f}_i\Vert_2 \big/50\right) \right]\\
&-\sum_{i=1}^{4}C_i^{\rm des}\left({\phi}_i\right)\left[1-\exp\left(-\Vert\boldsymbol{v}_{foot,i}^{xy}\Vert_2 \big/1.25\right) \right].
\end{aligned}
\end{multline}

The $r_t^c$ is dependent on the gait parameters explicitly, according to (\ref{leg phase compute}), (\ref{contact schedule}), and (\ref{contact-related reward}). This contact-related reward helps the robot to learn specific contact patterns based on user's commands and does not restrict the robot to specific foot trajectories. 
Combining the $r_t^s$ and $r_t^c$ could help the agent to learn the detailed control of multiple gaits with natural styles.
We also add a tracking cost of the target base height $h_b^{\rm cmd}$ to learn the control of the base height $h_b$ according to user's commands.

For the agents learning adaptive gait behaviors, the $h_b^{\rm cmd}$ is fixed at \SI{0.3}{m}, and we do not provide these agents with $r_t^s$ or $r_t^c$, which allows them to explore the latent space of gait skills without any gait prior.
The details of the reward functions are shown in Table \ref{tab:reward details}. 
The total reward $r_t$ is defined by:
\begin{equation} \label{total reward}
r_t =r^g_t+r^l_t+r^s_t+r^c_t.
\end{equation}

\begin{table}[htbp]
\setlength{\abovecaptionskip}{0.cm}
\setlength{\belowcaptionskip}{-0.cm}
\centering
\caption{Reward terms used for learning multiple gaits.}
\label{tab:reward details}
\begin{tabular}{@{}lll@{}}
\toprule

\textbf{Term}         & \textbf{Equation}       &\textbf{Weight}\\ \midrule
\multirow{2}{*}{$r^g$}& $\exp \left(-\Vert\boldsymbol{v}_{xy}^{\rm cmd}-\boldsymbol{v}_{xy}\Vert_2 \big/ 0.15 \right)$    & 1.0\\
                      & $\exp \left(-\Vert{\omega}_{z}^{\rm cmd}-{\omega}_{z}\Vert_2 \big/ 0.15 \right)$  & 0.5\\ \midrule
\multirow{5}{*}{$r^l$}& $-\Vert\boldsymbol{\tau}\Vert_2$     & $1\times10^{-4}$\\
                      & $-\Vert\boldsymbol{\ddot q}\Vert_2$      & $2.5\times10^{-7}$\\
                      & $-\Vert \boldsymbol{q}_{t-1}-\boldsymbol{q}_t\Vert_2$ & 0.1\\
                      & $-\Vert\boldsymbol{h}_{b}^{\rm cmd}-\boldsymbol{h}_{b}\Vert_2$ & 1.0\\
                      & $-n_{collision}$    & 0.1\\\midrule
$r^s$                 & (\ref{style reward})  & 0.5\\\midrule
$r^c$                 & (\ref{contact-related reward}) & 1.0\\ 
\bottomrule
\end{tabular}
\end{table}

\section{Training}
\subsection{Training  Setup}
We use IsaacGym simulator to train 4096 parallel agents on different types of terrains  \cite{rudin2022learning}. 
The overall training time was 32 hours of wall-clock time, using a single NVIDIA RTX 3090Ti GPU. Each RL episode lasts for a maximum time of \SI{20}{s} and terminates early if it reaches the termination criteria (trunk collisions with the ground). 
The control frequency of the policy is \SI{50}{\hertz} in the simulation.

To improve the robustness of our policy and facilitate policy transfer from simulation to the real world, we apply the same strategy for dynamics randomization as in our previous work \cite{wjzamp}.

\subsection{Training  Curriculum}\label{subsection: Gait Curriculum}
We split the total robots in half into two groups. 
One group is responsible for learning multiple common gaits on flat ground by using the gait encoder, named as common group. 
The other group is responsible for learning adaptive gait behaviors over challenging terrains by using the gait generator, named as adaptive group.

To learn multiple gait skills, we first design a gait curriculum for our robot. We provide five quadruped gaits for the common group by using different phase offsets ${\varphi}_{2,3,4}$, including walking, trotting, pacing, pronking, and bounding. 
The robots in the common group are randomly given one of the five gaits and a random ${\phi}_1$ for the front-left leg at the beginning of an episode, while the $f$, ${\phi}_{\rm stance}$, and $h_b$ are sampled independently and uniformly from $\left[\SI{1}{\hertz}, \SI{4}{\hertz}\right]$, $\left[0.25, 0.75\right]$, and $\left[\SI{0.1}{m}, \SI{0.4}{m}\right]$. 
The $\boldsymbol{c}_t$ for the common group is sampled according to the grid adaptive curriculum strategy \cite{margolis2022rapid} that extends the sampling range of $\boldsymbol{c}_t$ only if the tracking task of $\boldsymbol{c}_t$ is well performed using the specific gait. 
This gait curriculum can enable our robot to learn high-speed sprinting and spinning using multiple common gaits.

To learn adaptive gait behaviors, we adopt the same terrain curriculum used in our previous work for the adaptive group.  
At the beginning of the training, all robots in the adaptive group are equally assigned to five terrain types with the lowest difficulty, including rough flats, slopes, waves, stairs, and discrete steps. 
The robots are only moved to more difficult terrains once they have performed velocity commands tracking task well, otherwise they are reset to easier terrains. 
For the agents of the adaptive group on rough flat terrains, once they step out of the roughest flats, their $\boldsymbol{c}_t$ sampling schedules are changed to the grid adaptive curriculum strategy, in order to learn agile locomotion. More details can be found in \cite{wjzamp}.

To generate $\boldsymbol{z}_t$ for the low-level network during training, the common group uses only the gait encoder, while the adaptive group uses only the gait generator. 
These two groups of massively parallel agents concurrently construct the same latent space of gait skills, which helps our robot to learn the detailed control ability of multiple gaits and the adaptive motor ability of robust and agile locomotion.

\section{RESULTS AND DISCUSSION}
\textbf{Hardware:} Our controller is deployed on the Unitree Go1 Edu robot, which stands \SI{0.3}{\m} tall and weighs \SI{13}{\kilogram}. The sensors used on the robot consist of joint position encoders and an IMU. The trained  policy runs on the onboard computer Jetson TX2 NX, with a control frequency of \SI{50}{\hertz}. 

\subsection{Latent Space Analysis}
\begin{figure}[htbp]
\setlength{\abovecaptionskip}{0.cm}
\setlength{\belowcaptionskip}{-0.cm}
\centering
\includegraphics[width=\linewidth]{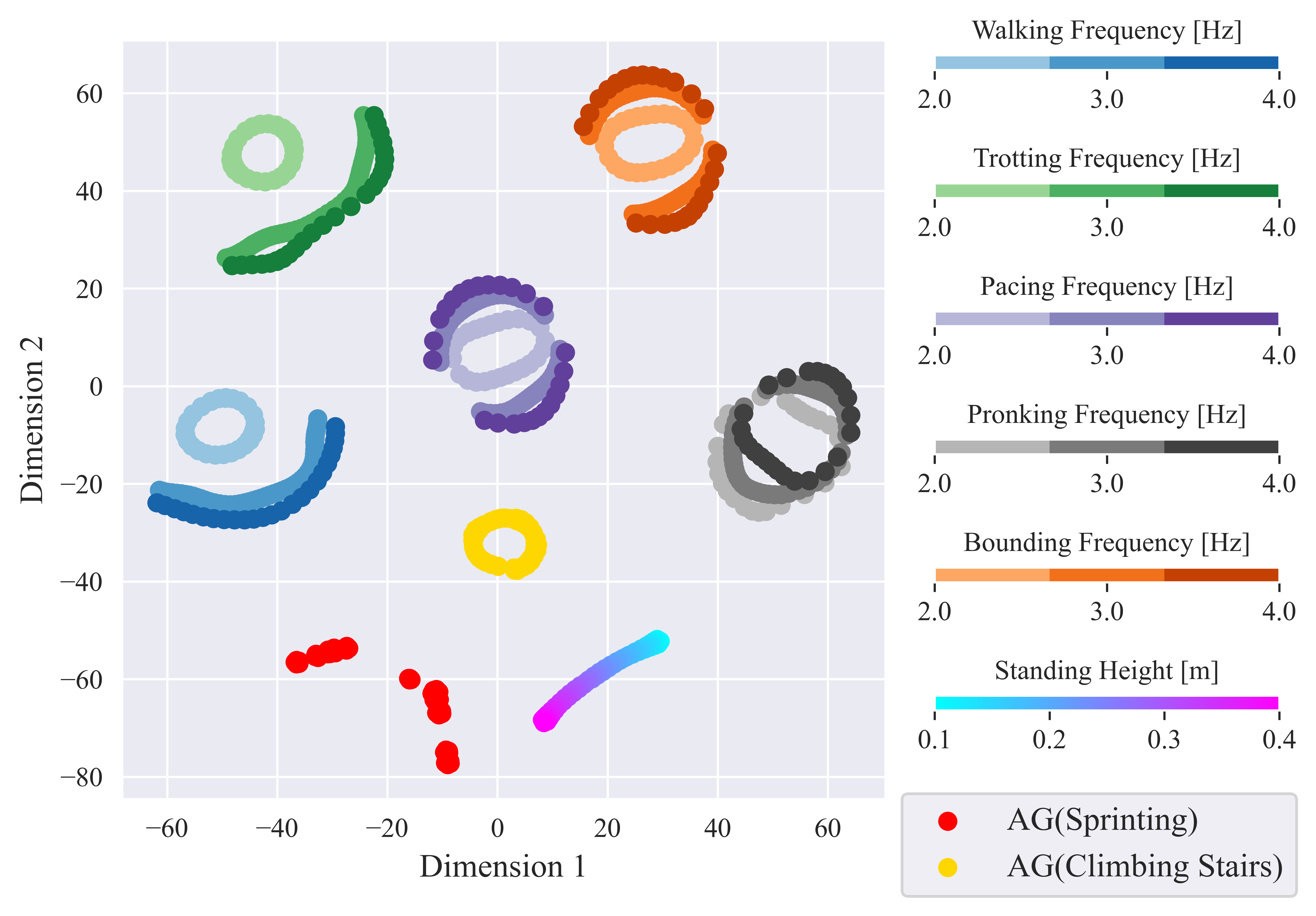}
\caption{The t-SNE visualization for the gait skills in the latent space.}
\label{fig:TSNE_space}
\end{figure}

To gain insight into the learned representations of gait skills, we analyzed the structure of the latent space. 
We used the gait encoder to encode six common gaits (walking, trotting, pacing, pronking, bounding, standing) and used the gait generator to generate adaptive gaits (AG) under two scenarios of agile and robust locomotion in simulation. 
Each locomotion gait in the common gaits had three $f$ of \SI{2}{\hertz}, \SI{3}{\hertz}, and \SI{4}{\hertz}, while the standing gait had a target base height increased from \SI{0.1}{m} to \SI{0,4}{m}, with a $f$ of \SI{0}{\hertz}. The first AG was generated for sprinting on the flat, while the second AG was generated for climbing stairs (\SI{25}{cm} width, \SI{20}{cm} height), given moving-forward commands of \SI{4}{m/s} and \SI{0.4}{m/s}, respectively. Each gait skill lasted 2 seconds.

We first conducted t-distributed stochastic neighbor embedding (t-SNE) on the learned latent representations for the above gait skills, as shown in Fig. \ref{fig:TSNE_space}. We found that the embedded latent representations were distributed distinctly for different gait behaviors and thus carried sufficient information about the gait skills. Moreover, the t-SNE plot shows that the latent representations of the same gait (different $f$) are clustered, which indicates that similar gait skills reside in proximity in the latent space.

\begin{figure}[htbp]
\setlength{\abovecaptionskip}{0.cm}
\setlength{\belowcaptionskip}{-0.cm}
\centering
\includegraphics[width=\linewidth]{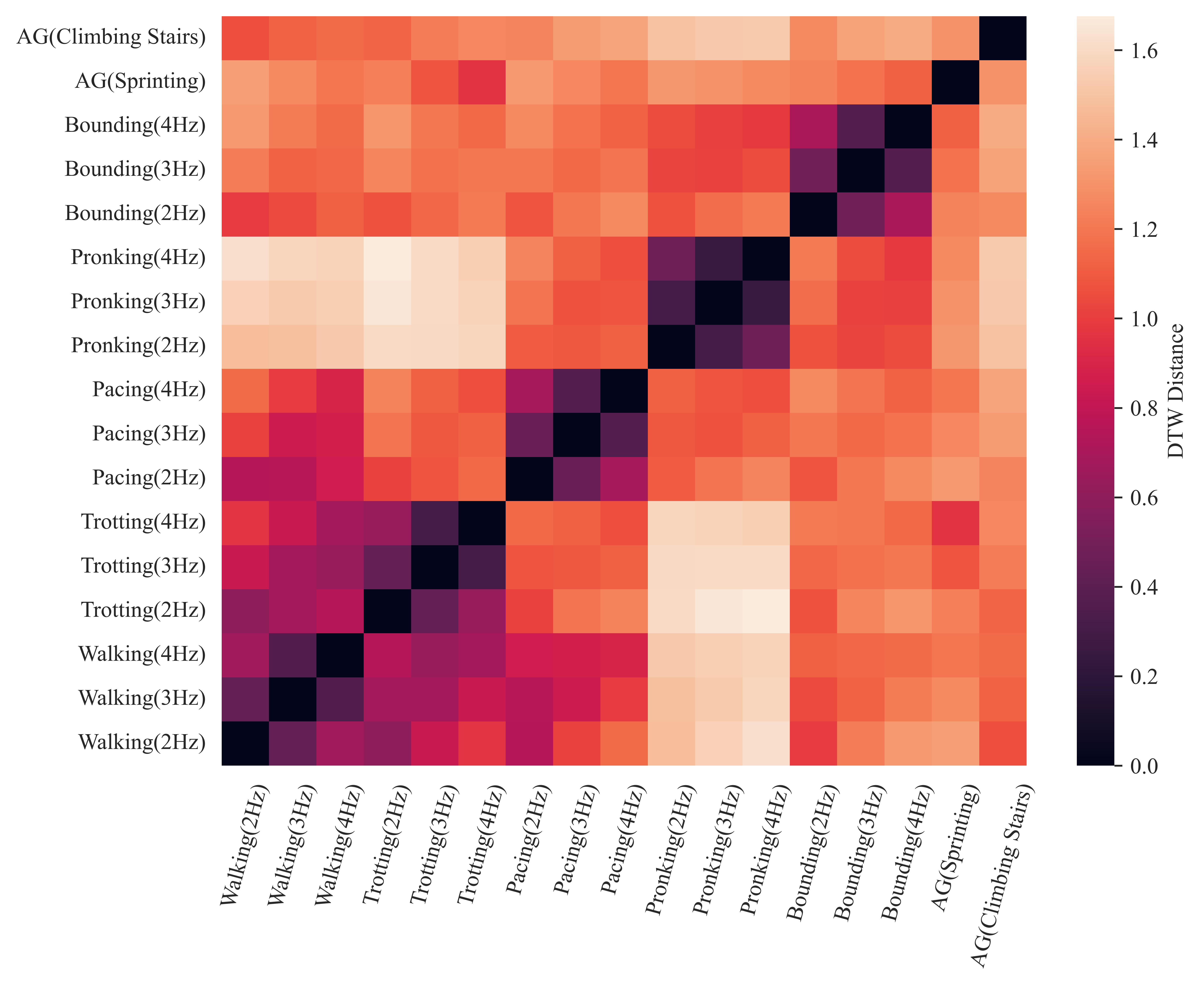}
\caption{The DTW distance of the gait skills in the latent space. Darker colors represent higher proximity between the two gait skills in the latent space.}
\label{fig:dtw_space}
\end{figure}

To measure the similarity and reusability between the gait skills, we calculated the distances between the latent representations of different gait skills using dynamic time warping (DTW). As shown in Fig. \ref{fig:dtw_space}, similar gait behaviors are closer in the latent space. This is observed by the lower values (darker) of the DTW distances along the diagonal region, which intuitively indicates that the distances between different frequencies of the same gait are closer than the distances between different gaits, in the latent space. 
The trotting and walking gaits are more similar to each other than to other gaits, while the pronking gait is more distinct from the other gaits. 

For the adaptive gait behaviors, the AG (sprinting) is similar to the trotting gait with high frequency (4Hz), while the AG (climbing stairs) is similar to the walking gait with low frequency (2Hz). 
The fact that the AG reuses these two gait skills is likely due to the different goals for different scenarios. 
For performing agile locomotion, the robot needs to maintain its balance and track velocity commands as much as possible, in order to gain high rewards during training. 
This incentive mechanism will cause the robot to select a gait skill to maintain stability and estimate its own speed as easily as possible. 
In this case, the trotting gait has advantages over other gaits, since it is more effective than the walking gait, more stable than the pacing gait, and does not have a distinct flying phase (pronking, bounding) that is difficult to estimate body velocities. 
For performing robust locomotion, the robot has to move as steadily as possible over challenging terrains. 
Since our policy uses only proprioceptive sensors, the robot needs to sense the external environment through multiple contacts between the foot and the terrain, in order to achieve blind locomotion. In this case, the walking gait is more stable than the other gaits.

\subsection{Ablation Study for the Learning Framework}
\begin{table*}[htbp]
\setlength{\abovecaptionskip}{0.cm}
\setlength{\belowcaptionskip}{-0.cm}
\centering
\caption{Ablation study for the concurrent learning framework: agile and robust tests.}
\label{tab:Ablation Study}
\begin{tabular}{@{}lllllllll@{}}
\toprule
\multirow{2}{*}{Terrain} & \multirow{2}{*}{Task parameters} & \multicolumn{7}{c}{Gaits}\\ \cmidrule(l){3-9} & 
& Adaptive(\textbf{C}) & Adaptive(H) & Walking & Trotting & Pacing & Pronking & Bounding \\ \midrule
\multicolumn{9}{c}{
Agile locomotion test for following longitudinal velocity commands (tracking error [RMSE])}
\\\midrule
\multirow{4}{*}{Normal flat} 
&$v_x^{\rm cmd}=\SI{1}{m/s}$&$\bm{0.042^{\pm0.005}}$
&$0.056^{\pm0.011}$&$0.115^{\pm0.010}$&$0.095^{\pm0.021}$
&$0.085^{\pm0.017}$&$0.085^{\pm0.032}$&$0.098^{\pm0.016}$ \\
&$v_x^{\rm cmd}=\SI{2}{m/s}$&$\bm{0.035^{\pm0.006}}$
&$0.078^{\pm0.025}$&$0.065^{\pm0.026}$&$0.064^{\pm0.026}$
&$0.104^{\pm0.031}$&$0.121^{\pm0.040}$&$0.104^{\pm0.044}$\\
&$v_x^{\rm cmd}=\SI{3}{m/s}$&$\bm{0.053^{\pm0.017}}$
&$0.169^{\pm0.077}$&$0.133^{\pm0.067}$&$0.129^{\pm0.068}$
&$0.249^{\pm0.058}$&$0.244^{\pm0.112}$&$0.231^{\pm0.090}$\\
&$v_x^{\rm cmd}=\SI{4}{m/s}$&$\bm{0.085^{\pm0.034}}$
&$0.128^{\pm0.051}$&$0.463^{\pm0.045}$&$0.509^{\pm0.051}$
&$1.427^{\pm0.698}$&$1.096^{\pm0.299}$&$1.474^{\pm0.479}$\\ \midrule
\multirow{4}{*}{Rough flat} 
&$v_x^{\rm cmd}=\SI{1}{m/s}$&$\bm{0.043^{\pm0.007}}$
&$0.063^{\pm0.013}$&$0.119^{\pm0.011}$&$0.107^{\pm0.029}$   
&$0.097^{\pm0.021}$&$0.108^{\pm0.048}$&$0.104^{\pm0.015}$   \\
&$v_x^{\rm cmd}=\SI{2}{m/s}$&$\bm{0.053^{\pm0.012}}$ 
&$0.077^{\pm0.020}$&$0.097^{\pm0.030}$&$0.146^{\pm0.081}$   
&$0.138^{\pm0.021}$&$0.179^{\pm0.032}$&$0.131^{\pm0.032}$   \\
&$v_x^{\rm cmd}=\SI{3}{m/s}$&$\bm{0.087^{\pm0.025}}$ 
&$0.127^{\pm0.050}$&$0.204^{\pm0.073}$&$0.479^{\pm0.454}$   
&$0.337^{\pm0.035}$&$0.493^{\pm0.325}$&$0.544^{\pm0.361}$   \\
&$v_x^{\rm cmd}=\SI{4}{m/s}$&$\bm{0.170^{\pm0.053}}$ 
&$0.338^{\pm0.185}$&$0.820^{\pm0.407}$&$0.823^{\pm0.409}$   
&$1.779^{\pm0.320}$&$1.312^{\pm0.364}$&$2.207^{\pm0.172}$   \\ \midrule
\multicolumn{9}{c}{
Robust locomotion test for climbing different terrains (success rate [$\%$])}
\\\midrule
\multirow{4}{*}{Stairs} 
&$h_{\rm stair}=\SI{5}{cm}$  &\textbf{100}&100& 0  & 0  & 0 & 0 & 20 \\
&$h_{\rm stair}=\SI{10}{cm}$ &\textbf{100}&60 & 0  & 0  & 0 & 0 & 0  \\
&$h_{\rm stair}=\SI{15}{cm}$ &\textbf{100}&0  & 0  & 0  & 0 & 0 & 0  \\
&$h_{\rm stair}=\SI{20}{cm}$ &\textbf{100}&0  & 0  & 0  & 0 & 0 & 0  \\ \midrule
\multirow{4}{*}{Slopes}  
&$\theta_{\rm slope}=\SI{10}{deg}$ &\textbf{100}& 100& 100 & 100 & 100  & 100 & 100\\
&$\theta_{\rm slope}=\SI{20}{deg}$ &\textbf{100}& 100& 100 & 100 & 100  & 100 & 100\\
&$\theta_{\rm slope}=\SI{30}{deg}$ &\textbf{100}& 100& 0   & 0   & 0    & 0   & 0  \\
&$\theta_{\rm slope}=\SI{40}{deg}$ &\textbf{80}&   20& 0   & 0   & 0    & 0   & 0  \\ 
\bottomrule
\end{tabular}
\end{table*}

Our work focuses on learning multiple gait skills and reusing them to achieve adaptive gait behaviors. This thought is similar to the hierarchical reinforcement learning (HRL) used in previous works \cite{peng2022ase,tessler2023calm}, where HRL often learns to use the frozen low-level skill network to perform high-level tasks. 
To evaluate the effectiveness of our concurrent learning framework for the gait encoder and gait generator, we additionally trained an ablated policy using HRL, where the gait generator was trained to exploit the fixed latent space constructed by the gait encoder in the first phase.  
For a fair comparison, both policies used the same curriculum strategy detailed in Section \ref{subsection: Gait Curriculum}, the same reward functions detailed in Section \ref{subsection: Reward Terms Design} and the same random seed.
We performed two tests (agile and robust) in simulation using different gait skills, consisting of the adaptive (C) gait learned from the concurrent learning framework, the adaptive (H) gait learned from HRL, and the five common gaits (walking, trotting, pacing, pronking, bounding). The result is shown in Table \ref{tab:Ablation Study}.

The first test is for agile locomotion. All gaits were used for sprinting on the normal and rough flats following $v_x^{\rm cmd}$ increased from \SI{1}{m/s} to \SI{4}{m/s}, where the common gaits had the same frequency of \SI{4}{Hz}.
The rough flat was constructed by adding \SI{\pm3}{cm} noise to the height of the normal flat.
We conducted five runs for each tracking test and computed the root mean square error (RMSE), where the superscript in Table \ref{tab:Ablation Study} reports the standard deviation across five random seeds.
As shown in Table \ref{tab:Ablation Study}, the adaptive gaits (C, H) have smaller tracking errors than the common gaits when tracking large $v_x^{\rm cmd}$ (4m/s) in the normal flat and all $v_x^{\rm cmd}$ in the rough flat. 
This performance advantage can be explained by the fact that the adaptive gaits are able to reuse the gait skills that estimate their velocities more easily than the common gaits with fixed frequencies. 
Moreover, the adaptive (C) gait obtains the smallest tracking errors of all $v_x^{\rm cmd}$ in both normal and rough flats, which indicates that our concurrent learning framework is more effective than HRL, as the latent space of the gait skills is constructed concurrently by the gait encoder and gait generator.

The second test is for robust locomotion. All gaits were used for climbing stairs and slopes following $v_x^{\rm cmd}$ of \SI{0.4}{m/s}, where the common gaits had the same frequency of \SI{2}{Hz}. 
The stairs had a fixed width of \SI{25}{cm} and a step height increased from \SI{5}{cm} to \SI{20}{cm}. The inclination of the slopes increased from \SI{10}{deg} to \SI{40}{deg}.
We conducted five runs for each climbing test and computed success rates. A test was successful if the robot climbed over \SI{4}{m} in less than \SI{20}{s}.
As shown in Table \ref{tab:Ablation Study}, the adaptive gaits (C, H) have better climbing ability than the common gaits over challenging terrains, since the common gaits were trained on flat and focused on the detailed control of gait behaviors. 
Moreover, the adaptive (C) gait is more robust than the adaptive (H) gait over higher stairs and steeper slopes.
This performance difference demonstrates that our concurrent learning framework can help the low-level network to learn to perform adaptive gait behaviors according to the latent space constructed by the gait generator, while the low-level network in HRL is fixed and cannot adapt to the gait generator in the second phase.

\subsection{Outdoor Experiments}
The results in this section can also be found in the accompanying video\footnote{The video is available in \href{https://youtu.be/MoFm6_JVNko}{https://youtu.be/MoFm6\_JVNko}}.

We first evaluated the multiple gaits control of our robot outdoors, including walking, trotting, pacing, pronking, bounding, and adaptive gait.
The robot can quickly switch to the corresponding gait with natural style, according to user's commands. 
Moreover, our robot can achieve smooth transitions between different gaits. This smooth performance can be attributed to the latent space, which provides a smooth space for selecting different gait skills and interpolating gait transitions between different gait skills. 

\begin{figure}[htbp]
\setlength{\abovecaptionskip}{0.cm}
\setlength{\belowcaptionskip}{-0.cm}
\centering
\includegraphics[width=\linewidth]{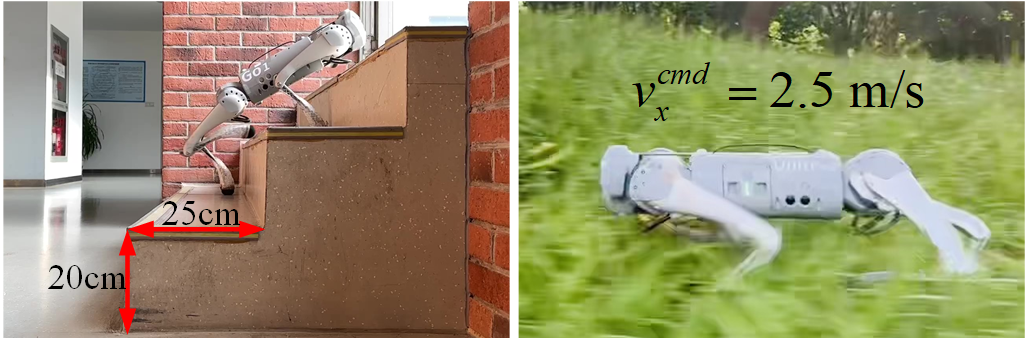}
\caption{The robust (left) and agile (right) tests outdoors.}
\label{fig:tests}
\end{figure}

We next compared the presented method with the following methods using only proprioception, in two tests of robust and agile locomotion:
\begin{enumerate}
\item \textbf{Baseline} \cite{wjzamp}: Our previous work used the teacher-student training framework and the AMP dataset of trotting gait.
\item \textbf{MoB} \cite{margolis2022walktheseways}: The policy was trained with more gait parameters, which can tune gait behaviors to aid generalization to different tasks, according to user's commands.
\end{enumerate}
The two tests are shown in Fig. \ref{fig:tests}. The robust test is climbing stairs with a width of \SI{25}{cm} and a height of \SI{20}{cm}, while the agile test is sprinting over vegetation with a $v_x^{\rm cmd}$ of \SI{2,5}{m/s}. 

We conducted five runs for each test and computed success rates. A test was successful if the robot could climb over the stairs or sprint out of the vegetation.
The result is shown in Table \ref{tab:cp outdoors}.
\begin{table}[htbp]
\setlength{\abovecaptionskip}{0.cm}
\setlength{\belowcaptionskip}{-0.cm}
\centering
\caption{Overall comparison with three methods.}
\label{tab:cp outdoors}
\begin{tabular}{@{}cccc@{}}
\toprule
Methods & Multiple gaits & Robust test ($\%$) & Agile test ($\%$) \\ \midrule
Baseline   & No             & 40                 & 60          \\
MoB        & Yes            & 0                  & 20           \\ 
Ours       & Yes            & 100                & 80           \\ \bottomrule
\end{tabular}
\end{table}
Since the Baseline was trained using only the AMP dataset of trotting gait, it is unable to achieve multiple gaits control compared to the MoB and Ours, which were trained with gait parameters and gait-dependent rewards. 
However, the Baseline and Ours are more robust and agile than the MoB, as shown in Table \ref{tab:cp outdoors}. This can be explained by the fact that the MoB was trained on the flat and focused on the detailed control of gait behaviors, while the others were trained over different terrains and had more flexibility to lean adaptive behaviors. 
Although the MoB can tune gait details to aid generalization to different tasks, it can hardly generalize to challenging terrains by tuning only gait parameters.
Moreover, Ours has higher success rates than the Baseline when performing robust and agile locomotion.
This performance difference is likely due to the different learning frameworks. 
Ours used the asymmetric actor-critic framework, which had smaller sim-to-real gap than the Baseline that used the teacher-student framework.

\section{CONCLUSIONS}
Our work demonstrates that multiple gait skills for quadruped robots can be generated using a single policy trained by RL, tailored to the needs of robust locomotion, agile locomotion, and user's commands. 
The latent space constructed concurrently by the gait encoder and gait generator can provide a smooth space for selecting different gait skills and interpolating gait transitions between different gait skills.

However, our system only uses proprioception and receives user's commands to achieve gait transitions.
In the future, our system could be extended with vision sensors, using exteroception to improve locomotion efficiency and switch gait automatically.

\bibliographystyle{./bibtex/IEEEtran} 
\normalem
\bibliography{./bibtex/my_refs}

\end{document}